\documentclass[conference,a4paper]{IEEEtran}
\IEEEoverridecommandlockouts
\usepackage{cite}
\usepackage{amsmath,amssymb,amsfonts}
\usepackage{algorithm}
\usepackage{algorithmic}
\usepackage{graphicx}
\usepackage{textcomp}
\usepackage{xcolor}
\usepackage{multirow}
\usepackage[caption=false]{subfig}
\usepackage{booktabs}
\usepackage[utf8]{inputenc}
\usepackage{flushend}
\def\BibTeX{{\rm B\kern-.05em{\sc i\kern-.025em b}\kern-.08em
    T\kern-.1667em\lower.7ex\hbox{E}\kern-.125emX}}
\begin{document}

\title{Efficient Learning of Model Weights via Changing Features During Training \\
%{\footnotesize \textsuperscript{*}Note: Sub-titles are not captured in Xplore and should not be used}
%\thanks{This work was supported in part by the project EFOP-3.6.2-16-2017-00015 supported by the European Union, co-financed by the European Social Fund.}
}

%\author{\IEEEauthorblockN{Ágnes Baran, and András Hajdu}
%\IEEEauthorblockA{\textit{Faculty of Informatics, University of Debrecen}\\
%Debrecen, Hungary \\
%e-mail: \{baran.agnes, hajdu.andras\}@inf.unideb.hu}
%}

\author{\IEEEauthorblockN{Marcell Beregi-Kovács}
\IEEEauthorblockA{\textit{Faculty of Mathematics}\\
\textit{University of Debrecen} \\
4002 Debrecen, POB 400, Hungary \\
beregi.kovacs.marcell@science.unideb.hu}
\and
\IEEEauthorblockN{Ágnes Baran}
\IEEEauthorblockA{\textit{Faculty of Informatics} \\
\textit{University of Debrecen}\\
4002 Debrecen, POB 400, Hungary \\
baran.agnes@inf.unideb.hu}
\and
\IEEEauthorblockN{András Hajdu}
\IEEEauthorblockA{\textit{Faculty of Informatics} \\
\textit{University of Debrecen}\\
4002 Debrecen, POB 400, Hungary \\
hajdu.andras@inf.unideb.hu}
}

\maketitle

\begin{abstract}
In this paper, we propose a machine learning model, which dynamically changes the features during training. Our main motivation is to update the model in a small content during the training process with replacing less descriptive features to new ones from a large pool. The main benefit is coming from the fact that opposite to the common practice we do not start training a new model from the scratch, but can keep the already learned weights. This procedure allows the scan of a large feature pool which together with keeping the complexity of the model leads to an increase of the model accuracy within the same training time. The efficiency of our approach is demonstrated in several classic machine learning scenarios including linear regression and neural network-based training. As a specific analysis towards signal processing, we have successfully tested our approach on the database MNIST for digit classification considering single pixel and pixel-pairs intensities as possible features.

\end{abstract}

\begin{IEEEkeywords}
machine learning, updating model weights, linear regression, neural networks, image classification
\end{IEEEkeywords}

\section{Introduction}
One of the most challenging problem of machine learning is to choose the proper features for training. A small number of features can result in underfitting, while using a too large feature set rises extremely the training time and increases the possibility of
overfitting, as well. 
For example in the case of regression it is well-known that a polynomial model usually  fits 
better the training data than a basic linear model. However, if we
have several features and would like to use a higher order polynomial
 model instead of the linear, we get many new features. In our method instead of working with  all the possible features we fix a specific number of them with using only a subset of the features corresponding to the polynomial model with a given order. 
During the training our aim is to find that subset of features which provides 
the best fit. This intention suggests the idea that during
the training starting from an initial subset we replace 
those features which are not significant with such ones that may
help to increase the accuracy. It means that we try many different models to find the best
combination of the features.
The basic idea of our approach is that the 
model is trained using a given number of iterations, then it eliminates
the ”useless” features and keeps only the ”useful” ones and it 
chooses new features instead of the eliminated ones. Those steps are repeated 
 until the stopping criterion fulfills  or all the features have been tested.

The main benefit of our approach comes from the fact that when the model changes during the training the weights of the remaining features can be kept. In other words, we do not perform a dedicated training for each different model which results in the saving of a remarkably amount of training time. 

Besides regression, our approach works similarly for classification scenarious. As for our practical example, we have examined the MNIST dataset and considered single pixel and pixel-pairs intensities as possible features.

There are a wide literature of the feature selection methods, see e.g. \cite{Stop}, \cite{Hua}, \cite{Siv}, \cite{Lai}, \cite{Zhang}
The methods can differ in the search strategies and in the criterion they used to calculate the relevance of the selected feature set. There exist methods which start from a minimal set of features and try to find the proper number and set of features (see \cite{Reun}, \cite{Zhao}), while other methods start from a complicated model and reduce the number of used features via a selection algorithm (see \cite{Oh}, \cite{Liu}, \cite{Ghe})

The works \cite{Kots}, \cite{Chan} provide a good overview of the selection methods. 

Besides the proposed new feature selection algorithm the novelty of our method is that in the case of image classification we filter the used pixel points, omitted the pixel positions which seem to be non-significant.

The rest of the paper is organized as follows. In  Section II we describe the algorithm both for regression and classification, in Section III we give three test examples to check the effectiveness of the proposed method, while in Section IV the experimental results are presented.  

\section{The algorithm}

\subsection{Regression}

As a starting problem we would like to approximate the function 
\[
Y:{\mathbb R}^n \to {\mathbb R}
\]
by a polynomial. 
Let be given the training set $X=\{ X^{(1)}, \dots ,X^{(M)}\}$, where  $X^{(i)}\in {\mathbb R}^n$, while the corresponding observations are denoted by $Y^{(i)}$, $i=1,\dots ,M$. 
In the case of  multivariate linear regression the coordinates of the vectors $X^{(i)}$ will be used as the  features and we determine the weight vector $\Theta\in{\mathbb R}^n$ which minimizes the mean squared error (MSE) loss 
\begin{equation}
L_{MSE}(\Theta)= \frac {1}{2M}\sum\limits_{i=1}^M \left( \Theta ^T X^{(i)}-Y^{(i)}\right) ^2. 
\end{equation}

In the first step of our algorithm we determine from the $n$ initial features used by the linear regression all the possible multinomial terms with degree at most $d$, where $d\in{\mathbb N}$ is an input parameter of the method. During the training that set of features will be used as a pool set (${\cal P}$). We would like to use  a fixed $K$ number of features  ($K\in{\mathbb N}$) to approximate the target $Y$. In what follows we will use the notation ${\cal F}$ for the set of actually used features. 

We start the algorithm with a randomly initialized set ${\cal F}$, then we train a linear regression model in $I$ iteration steps,  where $I\in{\mathbb N}$. After this training section the method evaluates all $f_i\in \cal F$ feature's "usefulness" by     
\begin{equation}
U_{f_i}=\frac{|L_{MSE}\big(\widehat{\Theta_{i}}\big)-L_{MSE}\big(\Theta\big)|}{|\Theta(i)|}, 
\end{equation}
where $\Theta (i)$ denotes the $i$th coordinate of the parameter vector $\Theta$, while $\widehat{\Theta_{i}}\in {\mathbb R}^n$ is the parameter vector whose elements are the same as  in  $\Theta$ except the $i$th one, which is equal to 0:  
\begin{equation}
\widehat{\Theta_{i}}(j)=\begin{cases}
\Theta(j), & \text{if } j\ne i\\
0, & \text{if } j=i
\end{cases} 
\end{equation}
$(j=1,\dots,n)$. The $U_{f_i}$ value measures the relative change of the cost if the $f_i$ feature is omitted from the model. The method use the $U_{f_i}$ values to rank the features and eliminates the $e\in \mathbb N$ least "useful"  features. After this, the method randomly chooses $e$ new features from $\cal P \setminus \cal F$ and updates the set ${\cal F}$. Let $\overline{\cal{F}}$ be the set of the eliminated features and ${\cal B}$ the set of the $m$ most useful features from $\overline{\cal F}$. During the selection  the algorithm basically prefers the features which were not chosen previously, however, the $m$ most useful previously eliminated features  have a chance to get into the set ${\cal F}$ again. 

In the remaining part of the algorithm we repeat the previously defined steps for the 
feature switch, however we choose the new features not from $\cal P \setminus \cal F$, but 
from the candidate set ${\cal C}=\cal P \setminus {\cal F}\setminus \overline{{\cal F}}\cup {\cal B}$. 

The number $e$ of the features we eliminate in the $k$th step  decreases to the values $e_1,\dots ,e_s$ as $k$ reaches $k_1,\dots ,k_s$, where the first sequence is a decreasing, while the second one is an increasing sequence of integers. 

The algorithm terminates when $|{\cal C}|\leq e$, where $|{\cal C}|$ denotes the cardinality of ${\cal C}$ 

\vskip .3cm 
An overview of the proposed method: 
\vskip .5cm 

\begin{algorithmic}
\STATE I. {\bf Input parameters}: 

$d$ (the maximal degree of the polynomial),

$K$ (the number of the used features), 

$I$ (the number of iterations in the training), 

$e$ (the number of the eliminated features),  

$m$ (the number of the most useful eliminated feature)

$(k_1,\dots ,k_s)$ (number of steps where  e is changing)

$(e_1,\dots ,e_s)$   (new values for e after $(k_1,\dots ,k_s)$ step)
\end{algorithmic}

\vskip .3cm 
\begin{algorithmic}
\STATE II. {\bf Initialization}

determine ${\cal P}$, initialize randomly ${\cal F}$ and $\Theta_{init}$, let  $\overline{\cal F}=\emptyset$, ${\cal B}=\emptyset$, ${\cal C}=\cal P \setminus {\cal F}\setminus \overline{{\cal F}}\cup{\cal B}$, $k=0$ 
\end{algorithmic}

\vskip .3cm 
\begin{algorithmic}
\STATE{\sc THE ALGORITHM}
\WHILE{$|C|\leq e$} 
    \STATE 1. starting from $\Theta_{init}$ train the network (determine $\Theta$) in $I$ iterations using the features from ${\cal F}$. 
    \STATE 2. Set $\Theta _{init}=\Theta$. 
    \STATE 3. determine the values $U_{f_i}$ for $f_i\in {\cal F}$
    \STATE 4. delete $e$ features from ${\cal F}$ corresponding to the least $U_{f_i}$ values
    \STATE 5. add these features to $\overline{\cal F}$
    \STATE 6. overwrite ${\cal B}$ with the set of the $m$ most useful features from $\overline{\cal F}$ 
    \STATE 7. ${\cal C}=\cal P \setminus {\cal F}\setminus \overline{{\cal F}}\cup{\cal B}$
    \STATE 8. choose randomly $e$ features from ${\cal C}$ and add these features to the set ${\cal F}$ 
    \IF{$k=k_i$} 
        \STATE $e=e_i$
    \ENDIF 
    \STATE $k=k+1$
\ENDWHILE 
\end{algorithmic}

\subsection{Classification}

We define a slightly modified version of the previous algorithm for classification with a multilayer perceptron (MLP). 

Due to the higher complexity of the problem we use an other function $U_{f_i}$ to measure 
the usefulness of the features. Let 
\[
U_{f_i}= \Vert \Theta ^{(f_i)} \Vert ,
\]
where $\Theta ^{(f_i)}$ denotes the column vector of the input layer's weight matrix corresponding to the feature $f_i$, while the norm is the Euclidean norm. 

The algorithm terminates if the accuracy on the training set becomes greater than a given value. In contrast to the algorithm given for regression  the number $e$ of the switched features  does not change during the training. We applied our method for image classification problems, the further details of the algorithm are written in the next sections.

\section{Test problems}
\subsection{First example}
In this example we use a basic linear regression to demonstrate effectiveness of the model. We generate ten features from Gaussian distribution with 10000 samples for training and 2000 samples for testing. Offline we calculate the all possible 5th degree multinomial term from the ten features. This gives 2992 new features plus the bias, so we get 3003 features. From these we randomly choose fifty features, its random linear combinations give the  function $Y$. For comparability we let our model and also the linear model to  learn for the same duration time (15000ms). Our model randomly gets fifty features and starts learning through fifty iterations by gradient descent and then it evaluates all feature's usefulness.  With this $U_{f_i}$ values the features can be sorted by usefulness. Those features will be "useless" which have low $U_{f_i}$ values. In the beginning of learning, during every feature switch we change the actual ten most "useless" features until we reach 100 feature switches. After that until we reach the 300th feature switch we change the actually five most "useless" features. Then we change only the worst feature until $|C|$ becomes less than 2. It helps to accelerate the procedure and to mitigate overfitting, because in the beginning most of the features are possibly not "useful". Sometimes a "useful" feature could be eliminated. It happens when the model already have a feature that is similar to this "useful" feature. For example parity of $x^3$ and $x^5$ functions are the same and their values are close to each other if $x\in(1-\varepsilon,1+\varepsilon)$ where $\varepsilon>0$ so one of it can act like the other one. The worse feature is in the model before the better one is join the model. In this case the worse one can suppress the good one. To avoid from eliminating permanently a "useful" feature, the best eliminated features remain in the candidate set. The learning ends when all the candidate features run out.

\subsection{Second example}
In this example we show the introduced method for classification on the database MNIST,  however,  we use a modified MNIST dataset. The difference between the original dataset and the used one is that this has only 5000 samples and the size of the pictures is reduced from $28\times28$ to $20\times20$. When we would like to use images for classification we have to deal with many features like pixel intensities.  

In the case of this dataset there are many pixels which have the same intensity for almost all the pictures, e.g. the pixels lying close to the border of the picture. In order to avoid to calculate with those irrelevant pixels we perform the following algorithm. 
After flattering the images to 400 dimensional vectors we compute the deviation of the 
pixel intensities corresponding to the different coordinates, i.e. we calculate 
the values 
\begin{equation}
s_j=\sqrt{\frac{1}{N-1}\sum\limits_{i=1}^n\left(I^{(i)}(j)-\overline{I(j)}\right)^2}, 
\end{equation}
where $N$ is the number of the training vectors, $I^{(i)}(j)$ denotes the $j$th coordinate 
of the $i$th training sample and 
\begin{equation}
\overline{I(j)}=\frac 1N \sum\limits_{i=1}^N I^{(i)}(j), 
\end{equation}
$j=1,\dots ,400$.
Then we sort the pixel intensities in descending order by their deviations and we keep only those that give the 99\% of the sum of the deviation. This reduce the original 400 pixel intensities to 287 and we use only this 287 values to create the new features, namely the pixel-pair intensities and the square of pixel intensities. We apply the previous filtering method on the new features, too, that means we omit the most irrelevant features,  and finally we get 287+32020 features. 

\begin{table}[!ht]
\centering
 \caption{Results for regression}
\label{tab:dev1}
\setlength\tabcolsep{3pt}
\renewcommand{\arraystretch}{1.5}{
\begin{tabular}{|c||c|c|}
\hline 
& Our model &Linear regression\\
\hline
\hline
Average Training Cost (MSE)&2.23&2.26\\
\hline
Average Testing Cost (MSE)&2.64&19.43\\
\hline
\end{tabular}}
\end{table}

Without that filtering we have 80600 features which  results in the slow down of the training. From the 5000 samples we choose 4500 for the training set and 500 for test set. The parameters of the applied neural network are the followings: input layer's size is 400, it has one hidden layer with 20 neurons, while the output layer's size is 10, corresponding to the number of classes. The initial features consist of the original 287 pixel intensities and we complete them with 113 features which are selected from the 32020 features randomly. We use the cross entropy as the cost function and apply an $L_2$-regularization with $\lambda =3.9$ regularization parameter. The length of the training periods is 20 iterations. In every elimination step of the method we eliminate 20 features, i.e. the value of $e$ is constant during the training. The method terminates when the  accuracy on the training set hits 96.5\%.

\subsection{Third example}
In this example we use our previously introduced method for classification on the MNIST dataset, too, with the same 32307 features. In this case we reduce the number of the neurons in the hidden layer from twenty to five. We also reduce the regularization parameter from 3.9 to 2 and we rise the stopping criterion from 96.5\% to 97\%. 

\section{Experimental results} 
We have tested our algorithm on a computer with the following specification: CPU Intel i9-9900k @4.6 GHz RAM 32 Gb DDR4 2666 MHz.

\begin{table}[t]
\centering
 \caption{Results for classification}
\label{tab:dev2}
\setlength\tabcolsep{3pt}
\renewcommand{\arraystretch}{1.5}{
\begin{tabular}{|c||c|c|}
\hline 
& Our model & MLP\\
\hline
\hline
Average Test Accuracy &93\%&93\%\\
\hline
Average Training Time (ms)&2400&3000\\
\hline
\end{tabular}}
\end{table}

In the first example we used the following parameter setting: $d=5$, $K=50$, $I=50$, $e=10$, $m=3$, $(e_1,e_2)=(5,1)$, $(k_1,k_2)=(100,300)$. 

\begin{figure}[t]
\centering
\includegraphics[width=3.4in]{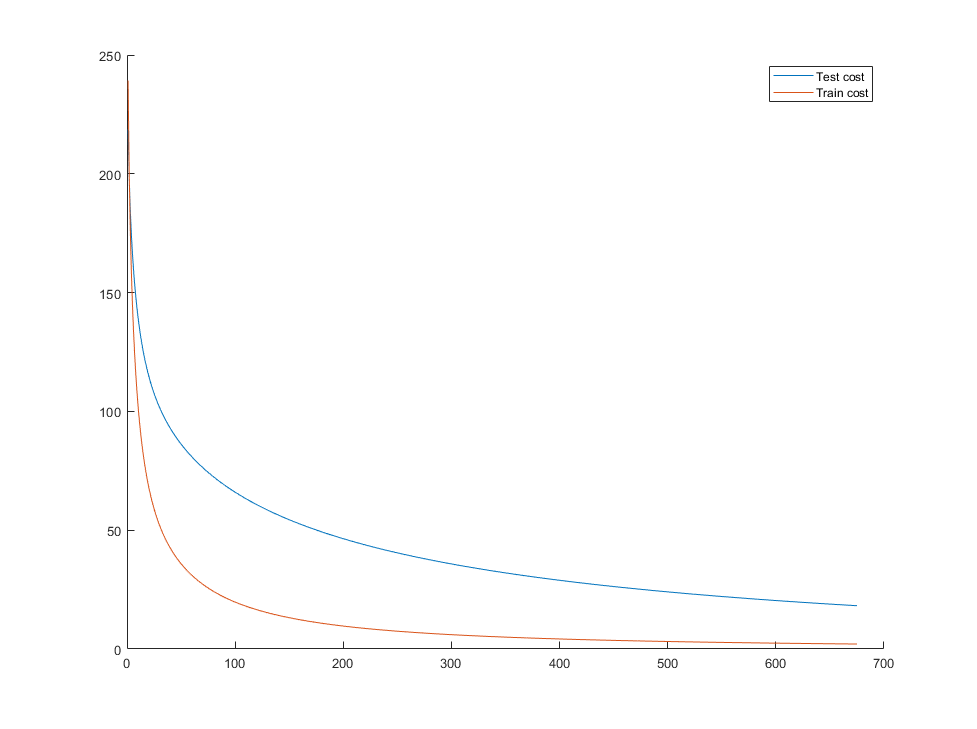}%
\caption{Cost function computed on the train and test sets for the basic linear regression}
\label{fig1}
\end{figure}

\begin{figure}[t]
\centering
\includegraphics[width=3.4in]{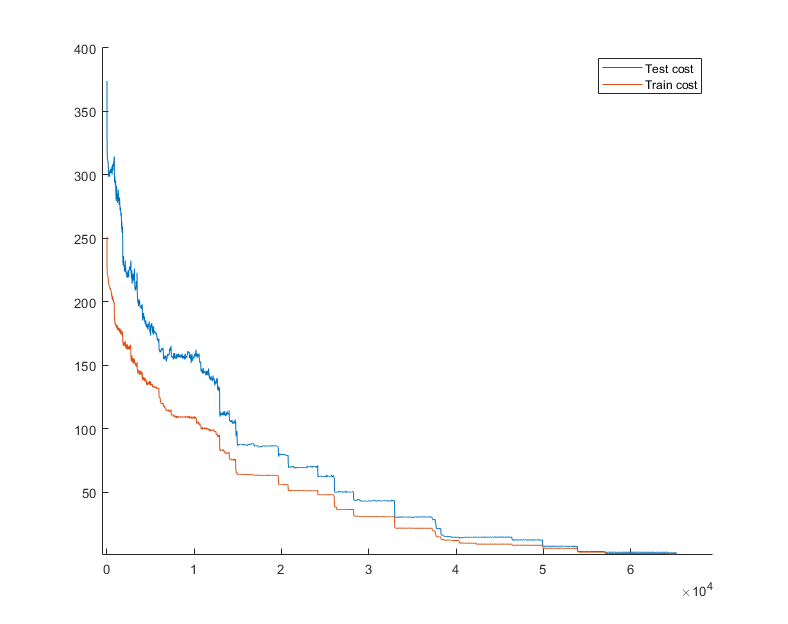}%
\caption{Cost function computed on the train and test sets for our method}
\label{fig2}
\end{figure}

Here, because we have lot of features (3003) the final model might be non perfect. (If we have much less feature the model can find the perfect combination of variables.) Meanwhile the basic linear regression model which gets all of the features and learns for the same time like our model (15000ms, 675 iteration) will be overfitted in the end of the learning and so it can not be used for prediction. 
The cost functions corresponding to the train and test sets for the basic linear regression and for our method  are plotted in Figure \ref{fig1} and \ref{fig2}, respectively. 
Otherwise the basic linear regression model uses much smaller learning rate because otherwise  the training will be divergent. The basic linear regression model uses $0.012$ learning-rate, while our model uses $0.1$. The results of both training are represented in the Table \ref{tab:dev1}.

The second method is compared with multilayer perceptron where the input layer's size corresponding to the size of the training images which is 400, it has one hidden layer with 20 neurons and the output layer's size is 10, i.e. size of the network is the same as the size of our network, however, it is trained only with the original pixel intensities.   The regularization parameter ($\lambda$) is 3 and this model is trained through 200 iterations (3000ms).  

In our algorithm the initial parameter settings were: $d=2$, $K=400$, $I=20$, $e=20$ (constant), $m=3$. 

We run the models with the previously mentioned settings 1000 times. Comparing the results both models reach 93\% average test accuracy, however for our method it takes only 2400ms averagely, see Table \ref{tab:dev2}.

\vskip .5cm 

In the case of the third example the initial values of the parameters of our network were  the same as before and we run the algorithm 1000 times. Because of the usage of a smaller network  the training time increased to 15500 ms averagely, while  the average test accuracy decreased to 91.17\%. The basic MLP has the same setting for the layers and it's regularization parameter is 1. The basic model is  trained on the original pixel intensities and it is trained for 1200 iterations (16000 ms). After the 1000 test runs we get 88.6\% test accuracy, see Table \ref{tab:dev5}.  

\begin{table}[t]
\centering
 \caption{Results for classification}
\label{tab:dev5}
\setlength\tabcolsep{3pt}
\renewcommand{\arraystretch}{1.5}{
\begin{tabular}{|c||c|c|}
\hline 
& Our model & MLP\\
\hline
\hline
Average Test Accuracy &91.17\%&88.6\%\\
\hline
Average Training Time (ms)&15500&16000\\
\hline
\end{tabular}}
\end{table}

\vskip .5cm

\section{Conclusion} 
In this paper we have presented a machine learning model which changes the features during the training in favour of finding a more useful set of features. We tested our method on image classification and linear regression tasks. With the help of our method we successfully reduced the training time or improved the test accuracy, so the test runs of our method show positive results. Therefore in the future we develop this method to come up with new results.

%\small
%\begin{thebibliography}{00}
\bibliographystyle{IEEEtran}
\bibliography{bibliography}
%\end{thebibliography}

\end{document}